\documentclass[runningheads]{llncs}
\usepackage[T1]{fontenc}
\usepackage{graphicx}
\usepackage{fancyvrb}
\usepackage{listings}
\usepackage[export]{adjustbox}
\usepackage[relative,overlay]{textpos}
\usepackage{tikz}
\usepackage{tabularx}
\usepackage{caption}
\usepackage{url}
\usepackage{colortbl}

\begin{document}
\title{Explaining Legal Concepts with Augmented Large Language Models (GPT-4)}
%
%
\author{Jaromir Savelka\inst{1}\orcidID{0000-0002-3674-5456} \and
Kevin D. Ashley\inst{2} \and\\
Morgan A. Gray\inst{2}\orcidID{0000-0002-3800-2103} \and\\
Hannes Westermann\inst{3}\orcidID{0000-0002-4527-7316} \and
Huihui Xu\inst{2}}
\authorrunning{J. Savelka et al.}
%
\institute{Computer Science Department, Carnegie Mellon University, Pittsburgh, USA \email{jsavelka@cs.cmu.edu}\and
Intelligent Systems Program, University of Pittsburgh, Pittsburgh, USA\and
Cyberjustice Laboratory, Faculté de droit, Université de Montréal, Québec, Canada}
\maketitle              
\begin{abstract}
Interpreting the meaning of legal open-textured terms is a key task of legal professionals. An important source for this interpretation is how the term was applied in previous court cases. In this paper, we evaluate the performance of GPT-4 in generating factually accurate, clear and relevant explanations of terms in legislation.
We compare the performance of a baseline setup, where GPT-4 is directly asked to explain a legal term, to an \textit{augmented} approach, where a legal information retrieval module is used to provide relevant context to the model, in the form of sentences from case law.
We found that the direct application of GPT-4 yields explanations that appear to be of very high quality on their surface. However, detailed analysis uncovered limitations in terms of the factual accuracy of the explanations. Further, we found that the augmentation leads to improved quality, and appears to eliminate the issue of hallucination, where models invent incorrect statements. These findings open the door to the building of systems that can autonomously retrieve relevant sentences from case law and condense them into a useful explanation for legal scholars, educators or practicing lawyers alike.

\keywords{GPT-4 \and Legal Concepts \and Augmented Large Language Models \and Legal Information Retrieval \and Statutory Provisions}
\end{abstract}
\section{Introduction}
This paper explores the ability of large language models (LLM), specifically GPT-4, to explain legal concepts from statutory provisions to legal professionals. We evaluate and compare the performance of the direct application of GPT-4 and an approach where the LLM is augmented with a legal information retrieval~(IR) component focused on retrieval of explanatory sentences from case-law. 

Legislative bodies enact statutes which are written laws that establish a set of legally enforceable rules. Typically, a statute addresses a particular subject and includes provisions that articulate individual legal principles, such as rights, restrictions, and obligations. Understanding statutory provisions can be challenging. This is because the provisions need to accommodate various circumstances, including those that have not yet occurred, within the abstract guidelines they convey. Hence, provisions of law communicate general standards and refer to classes of persons, acts, things, and circumstances \cite[p. 124]{Hart1994}. Consequently, legislators must use vague \cite{Endicott2000}, open textured \cite{Hart1994} terms, abstract standards \cite{Endicott2014}, principles, and values \cite{Daci2010} to deal with this uncertainty.

A key step in legal analysis is applying statutory provisions to factual situations. 
Deciding whether a specific legal term applies in a case requires the understanding of the precise meaning of such a term, through interpretation \cite{MacCormick1991}. Interpretation often centers around establishing the precise meaning of one or more words or phrases contained in the provision. Such an analysis typically includes referring to prior case-law, where courts have referred to, explained, interpreted or applied the statutory provisions \cite{savelka2022role}. Courts play a crucial role in interpreting the general statutory provisions in the light of specific situations, and may even be seen to change the law as they interpret it \cite{ashley2003law,vsavelka2015open}. Further, due to the concept of \textit{stare decisis}, courts are often bound to the precedent set by courts at the same or higher instance level. 

Thus, understanding how courts previously discussed a legal statutory term is an important area of investigation for interpreting statutory terms. In this work we investigate whether large language models, such as GPT-4, can be combined with existing legal information retrieval methods to automatically draft explanations of how previous courts explained the meaning of statutory terms. Such a system could be exposed to users on a website providing access to statutory laws, where clicking a term would bring up an explanation of the the term, based on case-law. Such a system could have an important impact in assisting lawyers in accomplishing their work more efficiently, or supporting judges in deciding how a term should be applied to a new situation. Eventually, it could also make the law more accessible for the public, although we here focus on explanations targeting an audience with legal training.


\begin{table}[th!]
  \footnotesize
  \centering
  \renewcommand{\arraystretch}{1.2}
  \begin{tabular}{|p{0.5\linewidth}|p{0.5\linewidth}|}
    \hline
    \textbf{Statutory Provision} & \textbf{Explanation (Short)} \\
    \hline
    46 U.S. Code \S\ 70001\newline Vessel traffic services\newline
    Subject to the requirements of section 70004, the Secretary—(3) (A) may require vessels to install and use specified \textcolor{blue}{navigation equipment}, communications equipment, electronic relative motion analyzer equipment, or any electronic or other device necessary to comply with a vessel traffic service or that is necessary in the interests of vessel safety. 
    & Courts have generally construed the term ``navigation equipment'' broadly to encompass various types of devices and systems that contribute to the safe and efficient navigation of vessels, including radar systems, electronic charting systems, and automatic identification systems [\ldots] (See United States v. Evergreen International S.A., 531 F.3d 321 (4th Cir. 2008); United States v. American Commercial Lines L.L.C., 759 F.3d 420 (5th Cir. 2014)). \\
    \hline
    \multicolumn{2}{|l|}{\textbf{Explanation (Long)}} \\
    \hline
    \multicolumn{2}{|p{\linewidth}|}{The term ``navigation equipment'' in the context of 46 U.S. Code \S\ 70001 has not been extensively litigated in the courts. However, the courts have addressed the broader issue of vessel safety and equipment requirements, acknowledging the importance of proper navigation equipment to ensure maritime safety.\newline
    Courts have recognized the importance of navigation equipment in maintaining maritime safety. In United States v. Evergreen Int'l Corp., 531 F. Supp. 2d 240 (E.D.N.Y. 2008), the court emphasized the need for vessels to properly install and use such equipment to prevent accidents and maintain safety on the high seas. [\ldots]\newline
    In other cases, courts have found that the failure to properly maintain or use navigation equipment can result in unseaworthiness claims. See, e.g., In Matter of Hechinger, 890 F.2d 202 (9th Cir. 1989) (holding that a vessel's lack of functioning navigation equipment rendered it unseaworthy).\newline
    In summary, while the term ``navigation equipment'' has not been extensively litigated in the context of 46 U.S. Code § 70001, courts have recognized the importance of proper navigation equipment in maintaining maritime safety and have upheld the Coast Guard's authority to regulate and enforce the use of such equipment. [\ldots]
    }\\
    \hline
  \end{tabular}
  \caption{Statutory provision with highlighted term of interest in the upper left, automatically generated short (1 sentence) explanation of its meaning in the upper right, and an automatically generated long (10 sentences) explanation in the bottom part (both abridged to fit into the table).}
  \label{table:task}
\end{table}

Table \ref{table:task} shows an example statutory provision focused on legal regulation of vessel traffic services (46 U.S. Code \S\ 70001). Supposing we are interested in the meaning of the phrase ``navigation equipment'' (highlighted blue in the table), the example explanations are shown in the upper right and bottom parts of Table~\ref{table:task}. Notice that the explanations are clearly meant for a legal professional as they use complicated language structures, legal terms, and references to legal sources. The explanations have been generated automatically by GPT-4 utilizing the text of the statutory provision, the information about the phrase the meaning of which should have been explained, and the prompt steering the language model to perform the desired task.

While the explanations appear reasonable there could potentially be many issues. For example, they could provide incomplete, inaccurate or even fictitious information. In this study, we investigate automatically generated explanations, such as the ones shown in Table \ref{table:task}. The main focus is on determining if the explanations can be improved by embedding the relevant information retrieved from case-law into GPT-4's prompt. Using this approach, GPT-4's role is mostly to organize the provided information into a concise coherent explanation as opposed to relying on GPT-4 to supply the needed information as well.

To investigate different techniques of generating explanations of legal concepts from statutory provisions, we analyzed the following research questions:

\begin{enumerate}
    \item What are the limitations of generating explanations directly with GPT-4?
    \item Does the quality of the explanations improve if the prompt provided to GPT-4 is augmented with relevant information retrieved from case-law and what are the properties of explanations generated in this way?
\end{enumerate}

\noindent By carrying out this work, we provide the following
contributions to the AI~\&~Law research community. As
far as we know, this is the first study that:

\begin{enumerate}
    \item Investigates the effectiveness of GPT-4 for explaining legal concepts in professional legal settings.
    \item Benchmarks the performance of a direct application of GPT-4 to the performance of GPT-4 augmented with legal information retrieval component.
\end{enumerate}

\section{Related Work}
The interpretation of open-textured legal terms has often come up in prior work, e.g. in the context of rule-based reasoning tools. Researchers have suggested providing case law extracts to the user to help them understand whether open-textured concepts apply, e.g. \cite{waterman1981models,paquin1991loge}. In \cite{westermann2023}, the researchers created summaries of how an open-textured concept was previously applied to support laypeople, and in \cite{salaun2022conditional}, the use of models to automatically summarize individual cases in this format was explored. In \cite{savelka2021legal}, the researchers examined the task of automatically retrieving sentences that are useful for the interpretation of a statutory term from case law. In \cite{savelka2021discovering}, pre-trained language models were used to automate the ranking of such sentences, outperforming previous approaches. Here, we use an augmented large language model to explain the insights gained from several previous cases selected using the methodology described in~\cite{savelka2021legal}. These explanations could give lawyers an overview of how a specific term was interpreted.

We use the GPT-4 model, combined with retrieved passages from court cases, to create explanations of the meaning of legal terms. There have been many investigations of GPT models in the legal domain, such as exploring whether they can be used to answer legal entailment questions in the Japanese bar exam \cite{https://doi.org/10.48550/arxiv.2212.01326}, to pass the Uniform Bar Examination \cite{katz2023gpt}, to conduct legal reasoning \cite{blair2023can,nguyen2023well}, to model U.S. supreme court cases \cite{hamilton2023blind}, to annotate legal documents \cite{savelka2023unlocking}, and to give legal information to laypeople \cite{tan2023}.

Several researchers have investigated the augmentation of language models, to retrieve more accurate and well-reasoned answers, or add new capabilities to the models \cite{mialon2023augmented}. This can be done through the structure of the prompt, i.e., by asking the models to show its reasoning steps (chain-of-thought or CoT) \cite{wei2022chain}, or other prompting methods.

Further, external information can be injected into the prompt given to the model, to guide its outputs. For example, sparse and dense retrieval methods can be used to identify documents that are of relevance to the specific user query, and can then be provided to the model as context \cite{mialon2023augmented,chen2017reading,asai2021one,lewis2020retrieval}. Language models can also take a more active role in retrieving the information \cite{schick2023toolformer}, e.g., by formulating a query for a web search engine \cite{thoppilan2022lamda,shuster2022blenderbot,yao2022react}, or even directly interacting with content and services on the internet \cite{nakano2021webgpt}. Here, we propose a pipeline to discover and insert relevant context into a prompt. Based on a user query for a specific legal term, we identify sections relevant for explaining a statutory term as in \cite{savelka2021legal}. These sections are then transformed by GPT-4 into a compact explanation.

\section{Proposed Method}
\label{sec:proposed_method}

In order to automatically create explanations of how legal terms have been previously interpreted, we use the GPT-4 model which is by far the most advanced model released by OpenAI as of the writing of this paper \cite{openai2023gpt4}. The model is focused on dialog between a user and a system (i.e., an assistant). An interesting property of models such as GPT-4 is that they appear to be very strong zero- and few-shot learners. This ability improves with the increasing size of the model~\cite{brown2020language}. The technical details of the GPT-4 model have not been disclosed~\cite{openai2023gpt4}.

\begin{figure}[t]
\footnotesize
\begin{Verbatim}[frame=single,commandchars=\\\{\}]
You are a legal assistant focused on statutory interpretation.
\end{Verbatim}
\begin{Verbatim}[frame=single,commandchars=\\\{\}]
Term of interest: \textcolor{blue}{\string{\string{term_of_interest\string}\string}}
Source Provision:
\textcolor{blue}{\string{\string{source_provision_citation\string}\string}}
\textcolor{blue}{\string{\string{source_provision_text\string}\string}}

\textcolor{orange}{From a legal information retrieval system, you receive a list of} 
\textcolor{orange}{sentences from case law mentioning a specific term of interest:}
\textcolor{orange}{- \textcolor{red}{\string{\string{sentence_1\string}\string}} (\textcolor{red}{\string{\string{case_citation_1\string}\string}})}
\textcolor{orange}{- \textcolor{red}{\string{\string{sentence_2\string}\string}} (\textcolor{red}{\string{\string{case_citation_2\string}\string}})}

For a user (legal professional), elaborate (in about \textcolor{blue}{\string{\string{num_sentences\string}\string}} 
sentence\textcolor{blue}{\string{\string{s\string}\string}}) on how the specific term of interest from a specified 
statutory provision has been explained or used by the courts in the 
past. Summarize the overall patterns present in the whole body of case-
law. Cite the most relevant court cases. In your explanations strictly 
adhere to the case-law. Focus on clarity. Provide a complete 
comprehensive explanation but do not stray from the topic.
\end{Verbatim}
\caption{GPT Prompts. The system prompt (top) provides high-level context. The task instructions are provided via the user prompt (bottom). Blue and red tokens with curly braces are replaced with the actual data. Orange and red passages are used only in the augmented version of the system.}
\label{fig:prompts}
\end{figure}

We set the \verb|temperature| of the model to 0.7, which is the default. The higher the \verb|temperature| the more creative the output but it can also be less factual. As the temperature approaches 0.0, the model becomes deterministic and can be repetitive. We set \verb|max_tokens| (i.e., the limit on the model's output) to 1,000 which is sufficient to accommodate the lengths of explanations we work with. We set \verb|top_p| to 1. This parameter is related to \verb|temperature| and also influences creativeness of the output. We set \verb|frequency_penalty| to 0, which allows repetition by ensuring no penalty is applied to repetitions. Finally, we set \verb|presence_penalty| to 0, ensuring no penalty is applied to tokens appearing multiple times in the output.

We compare the capability of two different model setups to determine the best way to generate explanations using such models: A baseline model and an augmented model.

\subsubsection{Baseline model}

Our baseline model works by directly asking GPT-4 to explain a term of interest, given the source provision and instructions to provide an explanation, i.e., without any further context. GPT-4 has been trained on an enormous corpus of documents. For reference, GPT-3, the previous model, was trained on a corpus of 570GB plain text files, including content crawled from the internet, books and Wikipedia \cite{brown2020language}. Thus, it is likely that the model was trained on many of the legal cases of higher instance courts, and may be able to use this information to explain the meaning of a term. On the other hand, models used in this setup may be prone to ``hallucination'', i.e., making up plausible sounding data. While GPT-4 is less likely than GPT-3 to hallucinate \cite{openai2023gpt4}, this remains an issue.

\begin{figure}[t]
\includegraphics[width=0.8\textwidth]{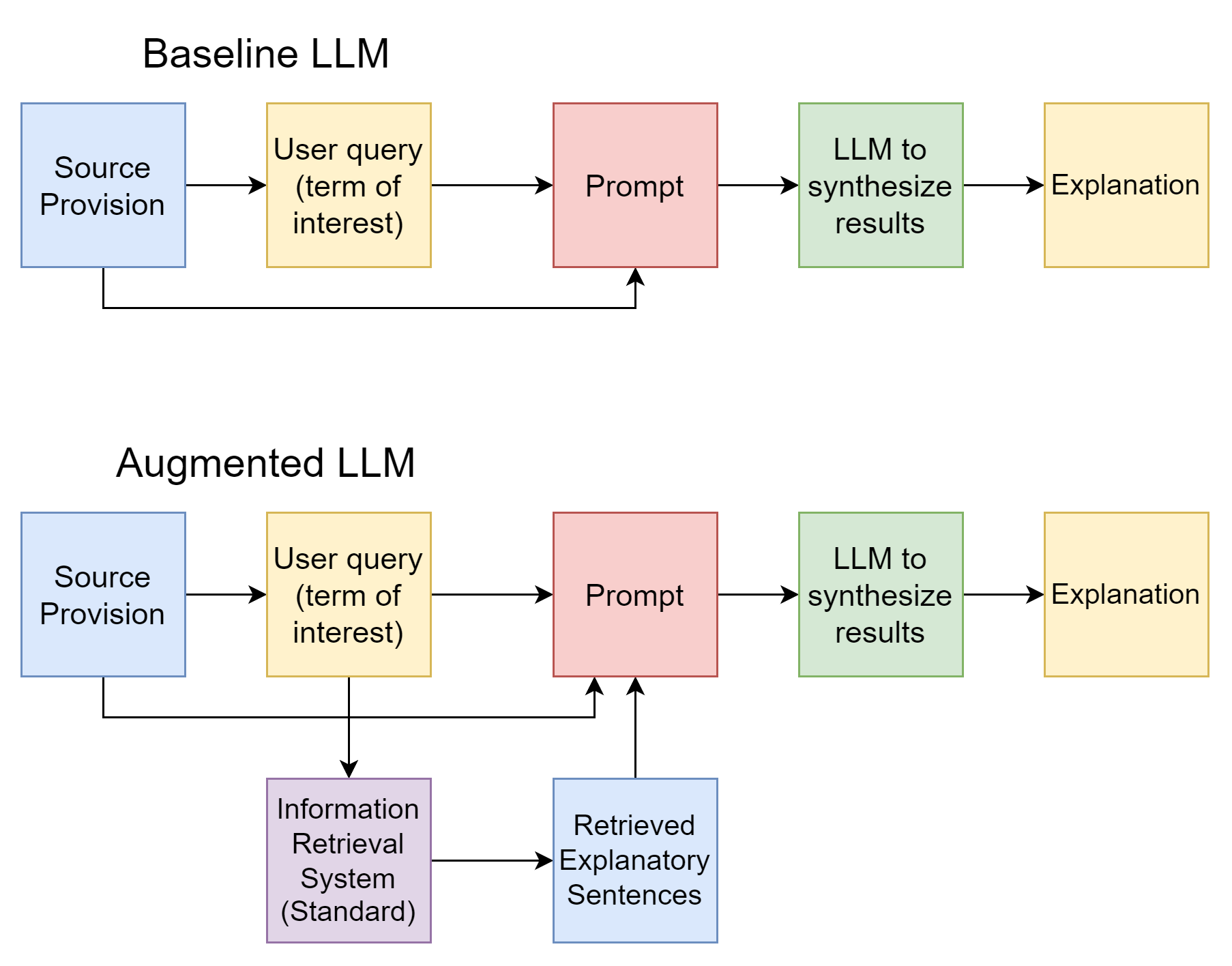}
\caption{System Architectures Diagrams. The top part shows the baseline directly applying the LLM. The bottom part describes the components of the augmented architecture that relying on the information retrieval component.}
\label{fig:system}
\end{figure}

The prompt used to steer GPT-4 toward the desired output is shown in Figure \ref{fig:prompts}. We use a so-called ``system prompt'', that tells the model that it is ``a legal assistant focused on statutory interpretation.'' Then, we provide the model with a term of interest, a source provision (i.e. the legal article containing the term), and instructions telling the model how long the summary should be, and providing some additional context on what the summary should focus on. The orange and red sentences in Figure \ref{fig:prompts} are not included in the baseline model prompt, i.e., all of the information the model outputs as a completion of the prompt is based on the data it has seen during training. The top part of Figure \ref{fig:system} shows the schematic diagram of the baseline model setup.


\subsubsection{Augmented model}
Next, we explore whether the GPT-4 model can be augmented by providing it with relevant sentences from case law that mention a term. Instead of relying solely on the training data to provide the context, here we add contextual data to the prompt of the model, which could reduce hallucination. GPT-4 is here not used as a search engine, but rather as a word calculator~\cite{Willison_2023}, aiming to transform the provided information into a compact explanation.

To evaluate this approach, we augment the prompt provided to GPT-4 with the ideal output of an information retrieval system proposed in \cite{savelka2021discovering}. For a fully working system, this would mean that we first retrieve full texts of court decisions responsive to a search query that corresponds to the term of interest (e.g., ``navigation equipment''). Then, the texts of the decisions are segmented into sentences \cite{savelka2017sentence}, and those that contain the term of interest are evaluated in terms of their value for explaining the term. The sentences that are deemed to have high value are then included in the prompt of the augmented system. The exact format of the prompt is shown in Figure \ref{fig:prompts}. Note that the difference between the prompt of the first version of the system (baseline) and the augmented version is the inclusion of the part highlighted in orange and red in Figure \ref{fig:prompts}.

The diagram describing the high-level architecture of the augmented system is shown in the bottom part of Figure \ref{fig:system}. The diagram shows the same inputs as in case of the baseline system, i.e., the term of interest and the source provision it comes from. However, the term of interest is also used to retrieve the cases in which it is mentioned, and from those the high value sentences are obtained. The list of these explanatory sentences is then included in the prompt alongside the term of interest and the source provision. The system outputs the explanation based on the information provided in the prompt.

\section{Experiments}

\subsection{Data}
To support the experiments in this work we use the publicly available statutory interpretation data set from \cite{savelka2021discovering,savelka2019improving,vsavelka2016extracting}.\footnote{Statutory Interpretation Data Set. Available at: \url{https://github.com/jsavelka/statutory_interpretation} [Accessed 2023-05-12]} The corpus contains 42 phrases from different statutory provisions of the United States Code. For each phrase the researchers collected a set of sentences mentioning the phrase from the court decisions retrieved from the Caselaw access project data.\footnote{Caselaw Access Project. Available at: \url{https://case.law/} [Accessed 2023-05-12]} In total the corpus consists of 26,959 sentences. The sentences were manually classified into four categories according to their usefulness for the interpretation---high value, certain value, potential value and no value. In this work we use the sentences labeled as high value only. These are sentences intended to define or elaborate on the meaning of the term. For example, high value sentences for the ``navigation equipment`` phrase from the example shown in Table \ref{table:task} could look like this:

\begin{quote}
The first subsection of that provision, entitled ``Navigation Equipment,'' requires tankers to possess global positioning system (``GPS'') receivers, as well as two separate radar systems.
\end{quote}

\noindent Overall, there are 1,853 high value sentences in the data set. The distribution across the phrases is quite diverse with the average of 43.1 high value sentences per phrase. The phrase associated with the most such sentences is ``fermented liquor'' (402 sentences in total), whereas the phrase with the least sentences is ``leadership role in an organization'' (0). These are the sentences that are used to augment the prompt provided to GPT-4.

\subsection{Experimental Design}
To investigate the two research questions, we first utilize the two variants of the system described in Section \ref{sec:proposed_method}, i.e., the baseline and the augmented model, to generate explanations for each of the 42 terms of interest. Each system generates two versions of an explanation for each term of interest---a short one (1 sentence) and a longer one (10 sentences). Hence, there are four explanations per term of interest (168 overall). The two explanations of the same type (i.e., either short or long) for the same term of interest were paired together for comparison performed by human annotators (84 pairs). We employed two human annotators who are both legal scholars with extensive experience in semantic annotation of legal texts (co-authors of this paper).

\begin{table}[t]
  \footnotesize
  \setlength{\tabcolsep}{12pt}
  \renewcommand{\arraystretch}{1.2}
  \centering
  \begin{tabular}{|l|l|l|l|}
    \hline
    \cellcolor{black!10}\textcolor{blue}{\{\{Term of interest\}\}} &\cellcolor{black!10} &\cellcolor{black!10} &\cellcolor{black!10} \\
    \hline
    \cellcolor{black!10}\textcolor{blue}{\{\{Source provision\}\}} &\cellcolor{black!10} &\cellcolor{black!10} &\cellcolor{black!10} \\
    \hline
    \cellcolor{black!10}  &\cellcolor{black!10}\textcolor{blue}{\{\{Explanation 1\}\}}&\cellcolor{black!10}\textcolor{blue}{\{\{Explanation 2\}\}}& \cellcolor{black!10}Notes\\ 
    \hline
    \cellcolor{black!10}Factuality           &\cellcolor{green!15} &\cellcolor{green!15} &\cellcolor{green!15} \\
    \hline
    \cellcolor{black!10}Clarity              &\cellcolor{green!15} &\cellcolor{green!15} &\cellcolor{green!15} \\
    \hline
    \cellcolor{black!10}Relevance            &\cellcolor{green!15} &\cellcolor{green!15} &\cellcolor{green!15} \\
    \hline
    \cellcolor{black!10}Information Richness &\cellcolor{green!15} &\cellcolor{green!15} &\cellcolor{green!15} \\
    \hline
    \cellcolor{black!10}On-pointedness       &\cellcolor{green!15} &\cellcolor{green!15} &\cellcolor{green!15} \\
    \hline
  \end{tabular}
  \caption{Annotation Task Form. The blue tokens with curly braces are replaced with the actual data. The task is to indicate preference if any for one of the explanations along the five observed dimensions. Optional comments could be provided in the Notes column.}
  \label{tab:annotation}
\end{table}

The overall layout of an annotation sheet for a comparison of a single pair of explanations is shown in Table \ref{tab:annotation}. Hence, the annotators were provided with the term of interest, the text of the source provision, as well as with the texts of the two explanations, each generated by a different version of the system. The annotators were not aware of the technical details behind the systems used to generate the explanations. The order in which the explanations were included in each sheet was randomized. However, the annotators knew that both of the explanations were generated automatically. The task of the annotators was to compare the explanations in terms of the following categories:

\begin{itemize}
    \item \emph{Factuality} -- The degree to which an explanation accurately represents the information, context, and judicial interpretations of the term of interest.
    \item \emph{Clarity} -- The degree to which an explanation is understandable, well organized, and effectively communicates the judicial interpretations of the term.
    \item \emph{Relevance} -- The degree to which an explanation is related to, and effectively addresses, the term by providing pertinent information from case-law.
    \item \emph{Information Richness} -- The degree to which an explanation provides a comprehensive, in-depth, and nuanced understanding of the term in question.
    \item \emph{On-pointedness} -- The degree to which an explanation remains focused on the term, its judicial interpretations, and its implications in the legal context.
\end{itemize}

\noindent The annotators were provided with detailed annotation guidelines that far exceeded the compact type definitions listed above. The goal was to decide which of the explanations is clearly better within each category, and mark it as such. The annotators also had an option of stating that none of the explanations was better within a category. Finally, the annotators were encouraged to comment on their decisions, although, this was not required (the Notes column in Table \ref{tab:annotation}). Each of the term of interest was handled by a single annotator, meaning both long and short explanation for a particular term of interest were evaluated by the same person.

\section{Results and Discussion}

\begin{figure}[t]
\includegraphics[width=\textwidth]{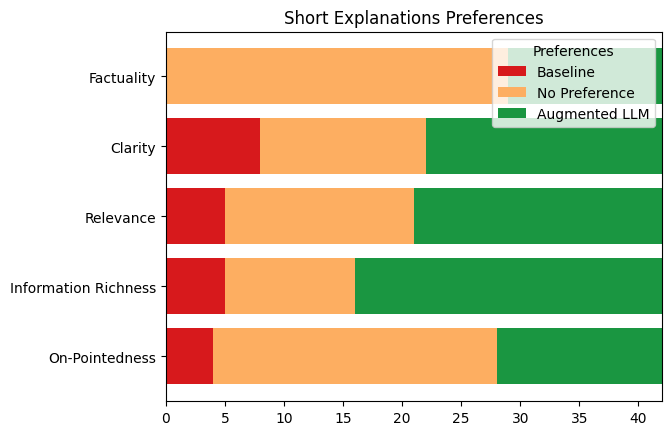}
\caption{Short Explanation Preferences. Red corresponds to the preferences for the explanations generated by the baseline system while green indicates preferences for the explanations coming from the augmented LLM. The yellow/orange informs about the number of instances where no preference was indicated.}
\label{fig:results_short}
\end{figure}

\begin{figure}[t]
\includegraphics[width=\textwidth]{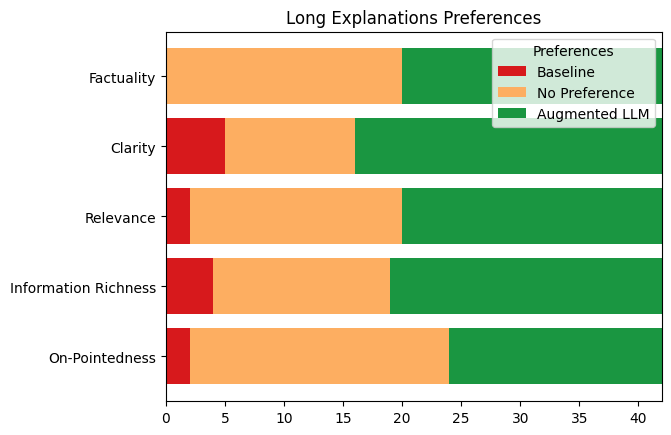}
\caption{Long Explanation Preferences. Red corresponds to the preferences for the explanations generated by the baseline system while green indicates preferences for the explanations coming from the augmented LLM. The yellow/orange informs about the number of instances where no preference was indicated.}
\label{fig:results_long}
\end{figure}

The results of the manual annotation effort comparing the short explanations generated by direct application of GPT-4 to those generated by the augmented system are reported in Figure \ref{fig:results_short}. It can be clearly seen that the short explanations generated by the augmented LLM are preferred over those that have been generated via direct application of GPT-4 without the access to the retrieved high value sentences. In certain cases, however, the annotators preferred explanations produced by the baseline system. Most often this happened in terms of clarity. On the other hand, there was not a single instance of preferring the explanation provided by the baseline system when it comes to factuality.

The results of the annotations for the long explanations, i.e., those that are about 10 sentences long, are reported in Figure \ref{fig:results_long}. Here, the preference for the explanations generated by the augmented LLM is even more pronounced than in the case of the short explanations. Otherwise, the overall patterns appear to be similar. The category in terms of which the explanations generated by the direct application of GPT-4 are preferred the most is clarity. Again, there was not a single case of preference for the explanation provided by the baseline system when assessed in terms of factuality.


\subsection{Factuality}
\subsubsection{Baseline}
For \emph{factuality} the annotators were asked to focus on accuracy, objectivity, and consistency. Since the explanations contained numerous citations to case-law one obvious concern was if the citations were real. This is because one of the well-known limitations of LLMs is the possibility of ``hallucination,'' i.e., stating of very plausibly sounding, yet, made-up information. Hence, the annotators were asked to confirm the existence of every cited case. Quite surprisingly, a large majority of citations generated by the direct application of GPT-4 were real existing citations. This is quite remarkable considering the citations typically look similar to the following: ``S.O.S., Inc. v. Payday, Inc., 886 F.2d 1081 (9th Cir. 1989).''
The fact that most of the time, GPT-4 does generate such citations without any mistakes in, e.g., numbering, is a testament to the incredible power of these LLMs to memorize data. However, the annotators identified multiple instances of citations where they were not successful in verifying the existence of the corresponding case. Further, even when the citation itself was correct, the information contained was often made up. Some cases were completely irrelevant, while for others the content was invented. This is a very serious limitation. A user needs to verify every single piece of information provided by the system to rule out the possibility of it being made up. This limitation alone requires legal professionals to be extremely vigilant when dealing with automatically generated explanations such as these. The identified issues pertain to both types of explanations--the short ones as well as the longer ones.



\subsubsection{Augmented LLM}
It appears the major issues identified in explanations generated by direct application of GPT-4 have been solved by augmenting the LLM, i.e., by inserting the list of high value sentences retrieved from case law into GPT-4's prompt. The annotators did not identify a single instance when an explanation generated by an augmented LLM cited a non-existing case or misrepresented what a cited case contains. Consequently, the annotators were more likely to prefer the augmented LLM compared to the baseline LLM.

However, several limitations were identified. These limitations include:
\begin{itemize}
    \item There were numerous citations to state cases while the term of interest comes from a federal statute.
    \item The information provided by the LLM is contained in the case, but is not useful or might even be misleading for the reader. In one instance, the system used a sentence from a dissenting opinion. In another instance, a sentence from a very old as well as overturned case was used.
    \item The term discussed in a case was the correct term, but in the case discussed in the context of a different law. 
\end{itemize}

\noindent All the above identified issues are attributable to the limitations of the legal IR component which supplied problematic results into the GPT-4's prompt. Hence, our findings emphasize the importance of the high quality of the augmentations to the LLMs. Any potential problems occurring in these components may propagate into the final result presented to users.

When evaluating the LLM in isolation, however, it does seem like augmenting the model with context in the form of retrieved sentences overcomes the issue of hallucinations, and thus presents a viable way to use LLMs to generate explanations of statutory terms.

\subsection{Clarity}
When evaluating the explanations on \emph{clarity} the annotators were asked to focus on language, structure, coherence, conciseness and consistency. Overall, the annotators were more likely to prefer the augmented LLM when it comes to clarity, for both the short explanations (Figure \ref{fig:results_short}) and the long explanations (Figure \ref{fig:results_long}). For the \textbf{baseline} model, a frequent comment on the clarity of the models was that the citations were formatted improperly, which was distracting in reading the summary. This was also the main comment regarding the \textbf{augmented LLM}. Interestingly, here, it seems like the incorrect formatting stems from the citations provided to the model in the prompt (see Figure \ref{fig:prompts}), which would be easily fixable. It is worth re-iterating that the annotators were legal professionals. Hence, the conclusion regarding clarity cannot be extended to laypeople---a separate study would be required for that.

\subsection{Relevance}
When evaluating the explanations on relevance the annotators were asked to focus on pertinence, importance, applicability and currency. The annotators preferred the explanations coming from the augmented system in terms of relevance, as evidenced by Figures \ref{fig:results_short} and \ref{fig:results_long}. There were no systematic issues discovered related to this evaluation category in the baseline model (beyond, of course, some of the cases not being factually correct, as discussed above). The annotators likewise did not provide a motivation for preferring the augmented LLM---a further investigation into differences between the two is left for future work.

\subsection{Information Richness}
When evaluating the explanations in terms of \emph{information richness} the annotators were asked to focus on depth, breadth, perspectives, contextual details, and nuance. Overall, the annotators had a preference for the explanations generated by the augmented LLM, as evidenced by Figures \ref{fig:results_short} and \ref{fig:results_long}. The annotators did not describe any specific issues with either of the models.

\subsection{On-pointedness}

Finally, when evaluating the explanations in terms of \emph{on-pointedness} the annotators were asked to evaluate the explanations in terms of their focus, precision and integration. It appears that the annotators decisively preferred the explanations coming from the augmented system as evidenced by Figures \ref{fig:results_short} and \ref{fig:results_long}. The annotators did not describe any specific issues with either of the models.

\section{Conclusions and Future Work}
We investigated the capacity of GPT-4 to explain legal concepts from statutory provisions. We found that while direct application of the LLM produced seemingly high-quality explanations at the surface level, certain limitations were discovered upon performing in-depth analysis. The most serious limitations were related to factuality of the produced explanations (e.g., citing non-existing case-law). The augmentation of GPT-4 with a legal information retrieval (IR) component significantly enhanced the explanations' quality across all the studied dimensions (e.g., factuality, clarity, on-pointedness), and overcomes the issues relating to hallucinations. This shows the potential for using LLMs coupled with traditional legal information retrieval methods to quickly produce accurate and useful summaries of how statutory terms have been interpreted. Such augmented LLMs could be valuable in teaching, learning, and practicing law, enhancing efficiency in workflows that require legal concept explanation and interpretation. 

Our work shows a promising way of overcoming one of the key limitations of LLMs, namely their propensity to hallucinate answers. The augmented LLM approach, that combines traditional methods in legal information retrieval with the power and flexibility of the language models, may thus be poised to play an important role in legal education and practice.

This research opens the door for different investigations for future work. First, we saw that shortcomings in the IR method used caused certain issues in the generated explanations. Overcoming these issues may further strenghten the generated explanations. Second, utilizing the approach to generate explanations for laypeople as in, e.g., \cite{westermann2023}, could be a valuable way to increase access to justice. Finally, using the augmented LLM approach with regard to other tasks in the legal domain could be a promising way to harness the power of LLMs.

\subsubsection{Acknowledgements} This work was supported in part by a National Institute of Justice Graduate Student Fellowship (Fellow: Jaromir Savelka) Award\# 2016-R2-CX-0010, “Recommendation System for Statutory Interpretation in Cybercrime,” a University of Pittsburgh Pitt Cyber Accelerator Grant entitled “Annotating Machine Learning Data for Interpreting Cyber-Crime Statutes,” and the National Science Foundation, grant no. 2040490, FAI: Using AI to Increase Fairness by Improving Access to Justice.

%
%
%
\bibliographystyle{splncs04}
%

\bibliography{citations}






\end{document}